\documentclass{article}

\usepackage[preprint]{neurips_2024}

\usepackage{graphicx}
\usepackage{cellspace}
\usepackage{subfigure}

\usepackage{float}
\usepackage{amsmath}
\usepackage{multirow}
\usepackage{color}
\usepackage{fontawesome}
\usepackage{algorithm}
\usepackage{algorithmic}

\usepackage[utf8]{inputenc} % allow utf-8 input
\usepackage[T1]{fontenc}    % use 8-bit T1 fonts
\usepackage{hyperref}       % hyperlinks
\usepackage{url}            % simple URL typesetting
\usepackage{booktabs}       % professional-quality tables
\usepackage{amsfonts}       % blackboard math symbols
\usepackage{nicefrac}       % compact symbols for 1/2, etc.
\usepackage{microtype}      % microtypography
\usepackage{xcolor}         % colors
\usepackage{makecell}
\usepackage{tikz}
\usepackage{overpic}

\title{Integrating Amortized Inference with Diffusion Models for Learning Clean Distribution from Corrupted Images}

\author{%
  Yifei Wang\thanks{Equal contribution.}\qquad Weimin Bai$^*$\qquad Weijian Luo\qquad Wenzheng Chen \qquad He Sun\thanks{Correspondence to hesun@pku.edu.cn} \\
  \\
  Peking University
}

\begin{document}

\maketitle

\begin{abstract}
% %% original 
% Diffusion models (DMs) have emerged as powerful generative methods for solving inverse problems, offering a good approximation of prior distributions of real-world image data. Typically, diffusion models rely on large-scale clean signals to sufficiently learn the score function of ground truth clean image distributions, a scenario often impractical in real-world applications due to the scarcity or unavailability of clean signals. To address this limitation, we introduce FlowDiff, a joint training paradigm that leverages a conditional normalizing flow to facilitate the training of diffusion models on corrupted data. The conditional normalizing flow estimates clean images for DM training through amortized inference. Our results show that FlowDiff can effectively learn clean distributions from various types of corrupted data, such as noisy and blurry images, and it consistently outperforms existing baselines under identical conditions. Additionally, the learned diffusion prior demonstrates superior performance in downstream computational imaging tasks.

%% new
Diffusion models (DMs) have emerged as powerful generative models for solving inverse problems, offering a good approximation of prior distributions of real-world image data. Typically, diffusion models rely on large-scale clean signals to accurately learn the score functions of ground truth clean image distributions. However, such a requirement for large amounts of clean data is often impractical in real-world applications, especially in fields where data samples are expensive to obtain. To address this limitation, in this work, we introduce \emph{FlowDiff}, a novel joint training paradigm that leverages a conditional normalizing flow model to facilitate the training of diffusion models on corrupted data sources. The conditional normalizing flow try to learn to recover clean images through a novel amortized inference mechanism, and can thus effectively facilitate the diffusion model's training with corrupted data. On the other side, diffusion models provide strong priors which in turn improve the quality of image recovery.
The flow model and the diffusion model can therefore promote each other and demonstrate %The FlowDiff has shown 
strong empirical performances. 
Our elaborate experiment shows that FlowDiff can effectively learn clean distributions across a wide range of corrupted data sources, such as noisy and blurry images. It consistently outperforms existing baselines with significant margins under identical conditions. Additionally, we also study the learned diffusion prior, observing its superior performance in downstream computational imaging tasks, including inpainting, denoising, and deblurring.

% , including denoising, deblurring, and inpainting.

% Diffusion models (DMs) have emerged as powerful generative solvers for inverse problems, owing to their impressive representation ability to capture complex distributions of real-world signals. 
% However, they are usually trained on large-scale clean signals to serve as plug-and-play priors, which is inconsistent with real-world cases where clean signals are expensive or even impossible to acquire. 
% To address this issue, we introduce FlowDiff, a joint training paradigm using both diffusion models and normalizing flow model following the Deep Probabilistic Imaging (DPI) framework. Using variational Bayesian methods, the diffusion model trained only with noisy, corrupted observations can learn a clean data distribution. 
% Besides the technically sound methods, empirical evidence shows that FlowDiff outperforms other baselines following the same setting where no clean signals are available. We demonstrate the utility of FlowDiff in downstream
% inference tasks such as denoising and deblurring.
\end{abstract}

\section{Introduction}
Diffusion models (DMs)~\cite{ho2020denoising,song2020score,song2019generative,sohl2015deep} have become a key focus in generative modeling due to their exceptional ability to capture complex data distributions and generate high-fidelity samples. Their versatility has led to successful applications across various data modalities, including images~\cite{rombach2022high,dhariwal2021diffusion}, videos~\cite{ho2022imagen,ho2022video}, text~\cite{li2022diffusion}, audio~\cite{kong2020diffwave}, 3D shapes~\cite{tang2023make,luo2021diffusion}, and scientific domains like molecule design~\cite{guo2024diffusion,huang2023mdm}. A particularly promising application of DMs is in solving computational imaging inverse problems, which aim to recover the underlying image $\mathbf{x}$ from noisy or corrupted observations $\mathbf{y}$~\cite{chung2022diffusion}. This can be probabilistically formulated as:
\begin{equation}
    p(\mathbf{x}\mid \mathbf{y}) \propto  p(\mathbf{y}\mid \mathbf{x})p(\mathbf{x}).
\label{equ:baye}
\end{equation}
where $p(\mathbf{y}\mid \mathbf{x})$ represents the forward model mapping images to observations, and $p(\mathbf{x})$ encodes prior knowledge about the images. Inverse problems are often ill-posed due to noise and corruption, leading to ambiguous solutions. DMs serve as powerful priors because they approximate the gradient of the data's log-likelihood function $\nabla_\mathbf{x} \log p_{data}(\mathbf{x})$, effectively constraining the solution space by leveraging their ability to model complex image distributions, favoring realistic and high-quality reconstructions.

However, training an effective DM typically requires a large dataset of clean images, which can be expensive or sometimes impossible to acquire. For example, in structural biology, 3D protein structures cannot be directly observed, and only low signal-to-noise 2D projections are captured by cryo-electron microscopy (Cryo-EM)~\cite{nogales2015cryo}. Similarly, in astronomy, black hole images are impossible to observe directly. Such scenarios, where only corrupted data is available, are common, especially in scientific applications. This raises the question: is it possible to train DMs on clean data distributions using only corrupted observations?

In this paper, we provide an affirmative answer and introduce a general framework capable of learning clean data distributions from arbitrary or mixed types of corrupted observations. The key insight is to incorporate an additional normalizing flow~\cite{kingma2018glow} that estimates clean images from corrupted observations through amortized inference. The normalizing flow is trained jointly with the DM in a variational inference framework: the normalizing flow generates clean images for training the DM, while the DM in turn imposes an image prior to guide the the normalizing flow model to get reasonable estimations % by the normalizing flow. 
Although this may seem like a chicken-and-egg problem, we demonstrate that a good equilibrium can be reached using an appropriate training strategy. The normalizing flow converges to a good inference function, and the DM converges to the clean data distribution, enabling simultaneous retrieval of clean images and learning of the clean distribution, even when no clean signals are provided. Through extensive experiments, we show that our method significantly outperforms existing approaches across multiple computational imaging applications, including denoising, deblurring, and fluorescent microscopy.

\section{Background}
\subsection{Score-based diffusion models}
Score-based diffusion models~\cite{ho2020denoising,song2020score,song2019generative,sohl2015deep} are a class of generative models that leverage stochastic processes to generate high-quality data, such as images or audio. Unlike traditional generative models that directly map latent codes to data samples, these models operate by gradually transforming a simple noise distribution, $\pi \sim \mathcal{N}(\mathbf{0},\mathbf{I})$, into a complex data distribution, $p_{data}$, through a series of small, iterative steps. This process involves a forward diffusion phase, which adds noise to the data, and a reverse diffusion phase, which denoises it. Both phases are governed by stochastic differential equations (SDEs) defined over the time interval $t \in [0, T]$:
\begin{equation} 
\begin{split} 
\text{Forward-time SDE:} \quad d\mathbf{x} &= \mathbf{f}(\mathbf{x},t)dt + g(t)d\mathbf{w}, \\ 
\text{Reverse-time SDE:} \quad d\mathbf{x} &= \left[\mathbf{f}(\mathbf{x},t) - g(t)^2\nabla_\mathbf{x} \log p_t(\mathbf{x})\right]dt + g(t)d\mathbf{\overline{w}}, 
\end{split} 
\label{eq:diffusionSDE}
\end{equation}
where $\mathbf{w} \in \mathbb{R}^d$ and $\mathbf{\overline{w}} \in \mathbb{R}^d$ are Brownian motions, $\mathbf{f}(\cdot, t): \mathbb{R}^d \rightarrow \mathbb{R}^d$ defines the drift coefficient that controls the deterministic evolution of $\mathbf{x}(t)$, and $g(\cdot): \mathbb{R} \rightarrow \mathbb{R}$ is the diffusion coefficient that controls the rate of noise increase in $\mathbf{x}(t)$.
At the core of the reverse diffusion process is a neural network, $s_\theta$, which is trained to approximate the score function, i.e., the gradient of the log-density $\nabla_\mathbf{x} \log p_t(\mathbf{x})$. This allows constructing a diffusion process $\mathbf{x}_{t=0:T}$ where $\mathbf{x}_0 \sim p_{data}$ and $\mathbf{x}_T \sim \mathcal{N}(\mathbf{0}, \mathbf{I})$. Score-based diffusion models have demonstrated remarkable success in generating high-fidelity data and have become a significant area of research in machine learning and artificial intelligence.

\subsection{Diffusion models for inverse problems}
Inverse problems aim to recover an underlying signal or image from observations and arise in various fields, such as computational imaging. These problems are often ill-posed due to factors like noise, incomplete data, or non-invertible forward operators ($\mathbf{x} \to \mathbf{y}$) that map the underlying signal $\mathbf{x}$ to observations $\mathbf{y}$. Traditional inverse problem solvers rely on handcrafted priors or regularizers that impose assumptions about the signal's structure or smoothness. However, these assumptions may lead to suboptimal reconstructions, especially for complex images with intricate details. Diffusion models offer powerful data-driven priors or regularizers for the inversion process~\cite{Luo2023ACS, feng2023score}. By incorporating the forward operator into the diffusion process using Bayes' rule, a conditional score function, $\nabla_\mathbf{x} \log p_t(\mathbf{x} \mid \mathbf{y})$, can be defined, enabling a conditional diffusion process to gradually recover the underlying clean image from noisy observations~\cite{graikos2022diffusion, kawar2021snips, chung2022diffusion}. As diffusion models inherently define a sampling approach, they not only provide point estimates but also quantify reconstruction uncertainties, which is valuable for scientific and medical imaging applications~\cite{song2021solving}.
% In recent years, diffusion models have found their use in solving inverse problems. Diffusion models can be used to reconstruct images from noisy or incomplete data. The trained diffusion model contains rich prior knowledge \cite{Luo2023ACS}. Therefore it can serve as strong prior distributions of ground truth data as demonstrated in \cite{feng2023score}. The model can sample from the real signal distribution conditioned on the noisy observations. A promising research track has been dedicated to refining the inference process to facilitate sampling from conditional distributions. ~\cite{graikos2022diffusion} illustrated that diffusion models can be effectively utilized as plug-and-play priors for image reconstruction. ~\cite{kawar2021snips} leveraged scored networks and enhanced sampling algorithms to address linear problems, while ~\cite{chung2022diffusion} extended these efforts to non-linear inverse problems. In a related context, ~\cite{song2021solving} pursued similar endeavors in the domain of medical image reconstruction. By acquiring knowledge of the true data distribution, these models can effectively recover clean images, conditioned on the noisy input data. However, it is essential to highlight that all these approaches presuppose that the diffusion model is trained on clean signals, which could be difficult and expensive to obtain, especially in the domain of medical imaging. Therefore, scenarios, where clean signals are impossible to acquire, is worthy to be considered. 

\subsection{Learning generative priors from corrupted data}
Generative models need large clean datasets to learn accurate data distributions. However, when only corrupted observations are available, directly training generative models on these data may lead to distorted or biased distributions, resulting in poor generative performance and suboptimal priors for inverse problems. A promising approach is learning the clean generative prior directly from corrupted observations. Early approaches like AmbientGAN~\cite{bora2018ambientgan} and AmbientFlow~\cite{kelkar2023ambientflow} have explored this concept for GANs and normalizing flows, respectively. AmbientGAN integrates the forward model into its generator, simulating the measurements of generated images, while its discriminator differentiates between real and simulated measurements. AmbientFlow uses a variational Bayesian framework~\cite{sun2021deep} to train two flow-based models; one predicts clean images from noisy data, and the other models the clean distribution. However, the limited capacity of GANs and flows restricts the modeling of complicated distributions

Recent research has shifted towards training clean diffusion models using corrupted data. Techniques such as AmbientDiffusion~\cite{daras2023ambient} introduce additional corruption during training. As the model cannot distinguish between original and further corruptions, this helps the diffusion model restore the clean distribution. Methods like SURE-Score~\cite{aali2023solving} and GSURE~\cite{kawar2023gsure} utilize Stein's Unbiased Risk Estimate (SURE) loss to jointly train denoising and diffusion models through denoising score matching. However, these methods often have restrictive assumptions about the type of corruption—SURE-Score is limited to denoising, and AmbientDiffusion to inpainting. This motivates exploring a more generalizable framework for training expressive diffusion models using arbitrary or mixed types of corrupted data. 

\section{Methods}
\begin{figure*}[tbp]
\centering
\includegraphics[width=14cm]{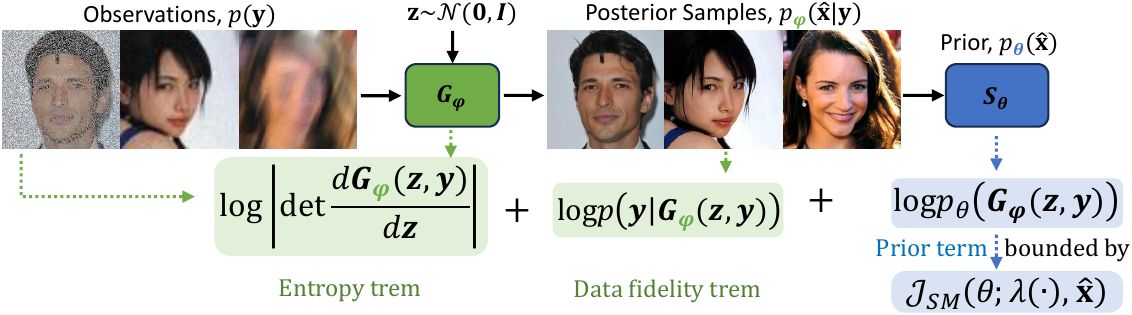}
\centering
\caption{\textbf{Overview of the FlowDiff.} We aim to train a clean diffusion model, $s_\theta$, using only corrupted observations. To achieve this, a conditional normalizing flow, $G_\varphi$, is introduced to recover underlying clean images through amortized inference. The conditional normalizing flow and the diffusion model are trained jointly: the flow generates clean images for training the diffusion model, while the diffusion model provides an image prior to regularize the output of the flow. Once the two networks reach equilibrium, clean reconstructions of corrupted observations are produced, and a clean diffusion prior is learned. }
\label{fig:overview}
\end{figure*}

In this section, we propose a diffusion-based framework to learn clean distribution from corrupted observations, as shown in Fig.~\ref{fig:overview}.
Specifically, we first introduce the amortized inference framework with a 
 parameterized normalizing flow for the image inverse problems in Sec.~\ref{sec:method-framework}.
Then, we elaborate on how to adopt score-based generative models to approximate image priors in Sec.~\ref{sec: method-calculatpriors}. 
Finally, we discuss the techniques and implementation details for jointly optimizing the normalizing flow and the diffusion prior in Sec.~\ref{sec:method-details}.

\subsection{Amortized inference with normalizing flows}
\label{sec:method-framework}
To solve a general noisy inverse problem $\mathbf{y}=f(\mathbf{x})+\eta$, where observations $\mathbf{y}$ are given, $f(\cdot )$ is the known forward model, $\eta \sim \mathcal{N}(\mathbf{0}, \sigma_n^2\mathbf{I})$, we aim to compute the posterior $p(\mathbf{x}\mid \mathbf{y})$ to recover underlying signals $\mathbf{x}$ from corrupted observations $\mathbf{y}$. 
However, it is intractable to compute the posterior exactly, as the true values of the measurement distribution $p(\mathbf{y})$ remain unknown. Therefore, we consider an amortized inference framework to approximate the underlying posterior with a deep neural network parameterized by $\varphi$. The goal of optimization is to minimize the Kullback-Leibler (KL) divergence between the variational distribution $p_\varphi(\mathbf{x}\mid \mathbf{y})$ and the true posterior $p(\mathbf{x}\mid \mathbf{y})$:
\begin{equation}
\begin{aligned}
    D_{KL}(p_\varphi(\mathbf{x}\mid \mathbf{y})\parallel p(\mathbf{x}\mid \mathbf{y}) )
    &= \int p_\varphi(\mathbf{x}\mid \mathbf{y}) \log \frac{p_\varphi(\mathbf{x}\mid \mathbf{y})}{p(\mathbf{x}\mid \mathbf{y})}d\mathbf{x} \\
    &= \int p_\varphi(\mathbf{x}\mid y) \log \frac{p_\varphi(\mathbf{x}\mid \mathbf{y})p(\mathbf{y})}{p(\mathbf{y}\mid \mathbf{x})p(\mathbf{x})}d\mathbf{x} \\
    &= \mathbb{E}_{p_\varphi(\mathbf{x}\mid \mathbf{y})} \left [ \log p_\varphi(\mathbf{x}\mid \mathbf{y}) + \log p(\mathbf{y}) - \log p(\mathbf{y}\mid \mathbf{x}) - \log p(\mathbf{x}) \right ]. 
\label{eq:objective-naive}
\end{aligned}
\end{equation}

Specifically, we introduce a conditional normalizing flow model $G_\varphi$ to
model the likelihood term $p_\varphi(\mathbf{x}\mid \mathbf{y})$.
Normalizing flows~\cite{dinh2014nice} are invertible generative models that can model complex distributions of target data~\cite{murphy2018machine}.
They draw samples $\mathbf{x}$ from a simple distribution(i.e. standard Gaussian distribution) $\pi(\mathbf{z})$ through a nonlinear but invertible transformation. The log-likelihood of samples from a normalizing flow can be analytically computed based on the “change of variables theorem”:
\begin{equation}
\begin{aligned}
    & \log p_\varphi(\mathbf{x}) = \log \pi(\mathbf{z}) - \log \left | \mathrm{det}\frac{dG_\varphi(\mathbf{z})}{d\mathbf{z}} \right |,
\label{eq:flow}
\end{aligned}
\end{equation}
where $\mathrm{det}\frac{dG_\varphi(\mathbf{z})}{d\mathbf{z}}$ is the determinant of the generative model’s Jacobian matrix.
Once the weights of $G_\varphi$ are trained, one can efficiently sample through the normalizing flow and exactly compute the log-likelihood of target samples.
These excellent properties make them natural tools for modeling the variational distribution. 
In the context of inverse problems, we adapt the unconditional flow to model the posterior distribution $p_\varphi(\mathbf{x}\mid \mathbf{y})$ conditioned on observations $\mathbf{y}$.
Therefore, we can further expand the KL divergence in Eq.~\ref{eq:objective-naive} based on conditional normalizing flows, $G_\varphi(\mathbf{z},\mathbf{y})$: 
% \wz{why we can direct change from $G_\varphi(\mathbf{z})$ to $G_\varphi(\mathbf{z},\mathbf{y})$  in the KL?}
\begin{equation}
\begin{aligned}
    \mathbb{E}_{\mathbf{z}\sim \pi(\mathbf{z})} \left [ \log \pi(\mathbf{z})-\underbrace{\log \left | \mathrm{det}\frac{dG_\varphi(\mathbf{z},\mathbf{y})}{d\mathbf{z}} \right |}_{L_{entropy}} + \log p(\mathbf{y})- \underbrace{\log p(\mathbf{y}\mid G_\varphi(\mathbf{z},\mathbf{y}))}_{L_{data fidelity}} - \underbrace{\log p(G_\varphi(\mathbf{z},\mathbf{y}))}_{L_{prior}} \right ].
    % + \mathbb{E}_{p_\varphi(\mathbf{x}\mid \mathbf{y})} \left[ - \log p(\mathbf{y}\mid G_\varphi(\mathbf{z},\mathbf{y})) - \log p(G_\varphi(\mathbf{z},\mathbf{y})) \right ]
\label{eq:objective-deduced}
\end{aligned}
\end{equation}
Since $\log \pi(\mathbf{z})$ and $\log p(\mathbf{y})$ are constant and have no learnable parameters, we simply ignore these terms during training. 
% As for the prior term $p(x)$, we simply plug in the ELBO of DDPM\ref{eq:ddpm-ELBO}. 
For simplification, we denote the posterior samples produced by the conditional normalizing flow, $G_\varphi(\mathbf{z},\mathbf{y})$, as $\hat{\mathbf{x}}$. The final objective function of the proposed amortized inference framework can be written as:
\begin{equation}
\begin{aligned}
    L = \mathbb{E}_{p(\hat{\mathbf{x}})} \left [ -\underbrace{\log \left | \mathrm{det}\frac{d\hat{\mathbf{x}}}{d\mathbf{z}} \right |}_{L_{entropy}} - \underbrace{\log p(\mathbf{y}\mid \hat{\mathbf{x}})}_{L_{data fidelity}} -  \underbrace{\log p(\hat{\mathbf{x}})}_{L_{prior}}\right ],
\label{eqn:objective-final}
\end{aligned}
\end{equation}
where the entropy loss $L_{entropy}$ and the data fidelity loss $L_{data fidelity}$ can be computed by the log-determinant of the generative model’s Jacobian matrix and data consistency, respectively.
As for the prior term, previous works mainly use handcrafted priors such as sparsity or total variation (TV)~\cite{kuramochi2018superresolution, bouman1993generalized}. 
However, these priors can not capture the complex nature of natural image distributions and always introduce human bias.
We propose to introduce data-driven DMs as the powerful priors.
Previous works are limited to simply adopting DMs pre-trained on a large dataset of clean images as priors~\cite{chung2022diffusion, feng2023efficient, feng2023score}, which can be expensive or sometimes impossible to acquire.  
Therefore, we carefully design an amortized inference framework to train the DM from scratch, without requiring any clean images.
% Notably, we train the DM from scratch, which is quite different from previous works that simply adopt pre-trained DMs for inference~\cite{chung2022diffusion, feng2023efficient, feng2023score}. 
% \wz{I would suggest to explain it a bit more.}

% where the prior term can be represented by its lower bound:
% \begin{equation}
%     \log p(\hat{x}) \ge -E_{q}[ \log p_{\theta}\left(\mathbf{x}_{0} \mid \mathbf{x}_{1}\right) ]+D_{\mathrm{KL}}\left(q\left(\mathbf{x}_{T} \mid \mathbf{x}_{0}\right) \| p\left(\mathbf{x}_{T}\right)\right) 
%     +\sum_{t>1} E_{q}[D_{\mathrm{KL}}\left(q\left(\mathbf{x}_{t-1} \mid \mathbf{x}_{t}, \mathbf{y}_{0}\right) \| p_{\theta}\left(\mathbf{x}_{t-1} \mid \mathbf{x}_{t}\right)\right)]
% \label{eq:objective-prior}
% \end{equation}

\subsection{Jointly optimizing score-based priors}
\label{sec: method-calculatpriors}
Ideally, we adopt a clean DM pre-trained on large-scale signals as a plug-and-play prior to Eq.~\ref{eqn:objective-final}. 
Extensive works have focused on solving general inverse problems in this paradigm. 
We argue that it is feasible to train this prior jointly with the amortized inference framework.
Assuming that the posterior samples are clean, we can train the score-based DM through well-established techniques.

\textbf{Maximum likelihood training of score-based diffusion models} 
Our goal is to learn the data distribution by approximating their score function $\nabla_\mathbf{x} \log p(\mathbf{x})$ with a neural network $\mathbf{s}_{\theta}$.
In order to train $s_{\theta}(\mathbf{x},t)$, ~\cite{song2019generative} proposed score matching loss
\begin{equation}
    \mathcal{J}_{SM}(\mathbf{\theta};\lambda(\cdot))=\frac{1}{2}\int_0^T\mathbb{E}_{p_t(\mathbf{x})}\left [\lambda(t)\Vert \nabla_{\mathbf{x}} \log p_t(\mathbf{x}) - \mathbf{s}_{\theta}(\mathbf{x},t)\Vert_2^2\right],
\label{eq:score-matching-loss}
\end{equation}
where $\lambda(t)$ is a weighting factor, i.e. $\lambda(t)=g(t)^2$.
Eq.~\ref{eq:score-matching-loss} stands for a weighted MSE loss between $\mathbf{s}_{\theta}(\mathbf{x},t)$ and $\nabla_\mathbf{x} \log p_t(\mathbf{x})$ with a manually chosen weighting function $\lambda(t)$. 
During training, we ignore $\lambda(t)$ for the benefits of sample quality (and simpler to implement)~\cite{ho2020denoising}.
Notably, the score matching loss could also serve as a data-driven regularizer during inference.

\textbf{Prior probability computed through DMs} 
By removing the Brownian motion from the reverse-time SDE in Eq.~\ref{eq:diffusionSDE}, that is, ignoring the stochastic term in the SDE, we can derive a probability flow ODE sampler~\cite{song2020score}:
\begin{equation}
    d\mathbf{x} = \left[ \mathbf{f}(\mathbf{x},t)-\frac{1}{2}g(t)^2\nabla_{\mathbf{x}} \log p_t(\mathbf{x})\right]dt.
\label{eq:ODE-sampler}
\end{equation}
~\cite{song2021maximum} proved that if $s_{\theta}(\mathbf{x},t) \equiv \nabla_\mathbf{x} \log p_t(\mathbf{x})$, Then $p_{\theta}^{\rm ODE}=p_{\theta}^{\rm SDE}=p_{data}$. Assuming the former equation always holds, we can compute the probability of a single image through integration:
\begin{equation}
\begin{aligned}
    \log p_{data}(\mathbf{x})&=\log p_{\theta}^{\rm ODE}(\mathbf{x}) \\
    &= \log \pi(\mathbf{x}(T))+\int_0^T \nabla_{\mathbf{x}} \cdot \left[ \mathbf{f}(\mathbf{x},t)-\frac{1}{2}g(t)^2\nabla_{\mathbf{x}} \log p_t(\mathbf{x})\right] dt.\\
    &= \log \pi(\mathbf{x}(T))+\int_0^T \nabla_{\mathbf{x}} \cdot \left[ \mathbf{f}(\mathbf{x},t)-\frac{1}{2}g(t)^2s_{\theta}(\mathbf{x},t)\right] dt.
\end{aligned}
\label{eq:ODE-probability}
\end{equation}
~\cite{feng2023score} has verified that a pre-trained $\mathbf{s}_{\theta}$ can serve as a powerful plug-and-play prior for inverse imaging.
However, the log-probability function in Eq.~\ref{eq:ODE-probability} is computationally expensive, requiring hundreds of discrete ODE time steps to accurately compute, thus not practical to use in the training process.
To reduce the computational overhead, ~\cite{song2021maximum, feng2023efficient} prove that the evidence lower bound (ELBO) could approximate the performance of $p_{\theta}^{\rm ODE}$ to serve as a prior distribution.
% Theoretically, with an ideal pre-trained DM, the image distribution derived from $p_{\theta}^{\rm ODE}$ is the same as that from $p_{\theta}^{\rm SDE}$~\cite{song2021maximum, feng2023efficient}. 
Interestingly, we find that in our setting, the score matching loss in Eq.~\ref{eq:score-matching-loss} becomes a lower bound of $\log p(\mathbf{x})$ by selecting a specific weighting function $\lambda(\cdot)=g(\cdot)^2$, which gives us an explicit function to approximate the probability of a prior image~\cite{song2021maximum}:
\begin{equation}
    -\mathbb{E}_{p(\mathbf{x})} \left [ \log p_{data} (\mathbf{x})\right]=-\mathbb{E}_{p(\mathbf{x})} \left [ \log p_{\theta}^{\text{SDE}} (\mathbf{x})\right] \leq \mathcal{J}_{SM}(\mathbf{\theta};g(\cdot)^2)+ C
\label{eq:sdebound}
\end{equation}
Consequently, by replacing $\mathbb{E}_{p(\hat{\mathbf{x}})} \left [- \log p(\hat{\mathbf{x}})\right ]$ in Eq.~\ref{eqn:objective-final} with the upper bound in Eq.~\ref{eq:sdebound}, we derive the loss for jointly optimizing the normalizing flow and the score-based diffusion prior.

% \textbf{Links between the score matching loss and ELBO} 
% We adopt the ELBO from denoising diffusion probabilistic models (DDPM)~\cite{ho2020denoising} for the simplicity of the problem, which can be seen as a discrete version of the former continuous-time equations~\cite{song2021maximum,huang2021variational}. 
% The ELBO can be formulated as:
% \begin{equation}
%     \begin{aligned}
%         \log p_\theta(\mathbf{x}) \geq&-E_{q(x_1|x_0)}[ \log p_{\theta}\left(\mathbf{x}_{0} \mid \mathbf{x}_{1}\right) ]+D_{\mathrm{KL}}\left(q\left(\mathbf{x}_{T} \mid \mathbf{x}_{0}\right) \| p\left(\mathbf{x}_{T}\right)\right) \\
%         &+\sum_{t>1} E_{q(x_1|x_0)}[D_{\mathrm{KL}}\left(q\left(\mathbf{x}_{t-1} \mid \mathbf{x}_{t}, \mathbf{y}_{0}\right) \| p_{\theta}\left(\mathbf{x}_{t-1} \mid \mathbf{x}_{t}\right)\right)]
%     \end{aligned}
%     \label{eq:ddpm-ELBO}
% \end{equation}
% where $0,1,..., T$ represents the discretized timesteps, $q$ represents the forward process and p equals the reversed process. 
% Given an image $\mathbf{x}$, we first randomly select a timestep $t$ and inject Gaussian noise into the picture through the forward-time SDE, deriving $q(x_t|x_0)$ and $q(x_{t-1}|x_t, x_0)$ can be explicitly represent by gaussian distributions. After that, we use a trained diffusion to model $p(x_{t}|x_0)$. At last, we can calculate ELBO.
% \textcolor{red}{Rewrite this paragraph.}

\subsection{Implementation details}
\label{sec:method-details}
% \wz{It may assume the readers are very familiar with ambient flow, ambient diffusion, I am not sure it is the case...But better to move the net architecture here with some more explanation, not just saying I use ambinetflow net arch}

\paragraph{Training scheme} 
We briefly discuss how to jointly train the normalizing flow and the diffusion model with the objective function in Eq.~\ref{eqn:objective-final}. Figure~\ref{fig:overview} illustrates our training framework. Our training loss consists of three terms: the first two terms (highlighted in green) are influenced only by the amortized inference network (i.e., conditional normalizing flow), while the third term (highlighted in blue) is influenced by both the inference network and the diffusion prior. In our joint optimization implementation, we alternate between updating the weights of the flow and the diffusion model in each training step. The flow is trained with all three terms, as defined in Eq.\ref{eqn:objective-final}, assuming the diffusion prior is fixed. Then, we fix the flow's weights and use the posterior images sampled from the flow model to train the DM using only the third term, i.e., the score matching loss defined in Eq.\ref{eq:score-matching-loss}. Note that when training the DM, we remove the weighting function $\lambda(t)$ in Eq.\ref{eq:score-matching-loss} for better performance, as suggested by Kingma et al.~\cite{kingma2024understanding}.

\paragraph{Model reset} 
Considering the joint optimization of the normalizing flow and diffusion model is highly non-convex, we often observe unstable model performance during training. Additionally, because both the normalizing flow and the diffusion prior are randomly initialized, they are initially trained with poor posterior samples, hindering optimal convergence due to memorization effects. To help the networks escape local minima, we periodically reset the weights of the normalizing flow and diffusion prior after a certain number of joint training steps.

For example, in denoising tasks, we reset the weights of the normalizing flow (amortized inference network) after 9000 joint training steps and retrain it from scratch until convergence, fixing the learned diffusion prior at the 9000th step. We then reverse the process: using the improved normalizing flow to generate better posterior samples, we retrain the diffusion model from scratch until it converges. Empirically, this model resetting strategy significantly reduces the influence of memorization effects, thereby enhancing both the amortized inference accuracy and the generative performance of the diffusion model. We also employ a similar strategy for the deblurring task.

% Considering the joint optimization of normalizing flow and DM is highly non-convex, we often observe a decrease in the model performance as the training goes on. However, fixing one of the model's weights and training the other model from scratch can ease this issue. For denoising tasks, We empirically observed that fixing the weights of the posterior model after 9000 training steps can enhance the performance of the DM. With the improved DM, we fix it weights and train the posterior model from scratch. Then we select the best checkpoint to train the DM from scratch until convergence. As for blurred images as inputs, we fix the posterior model after 9000 steps and train the DM from scratch as the final prior model directly. We remove the alternative training process for the posterior model can already produce deblurred outputs after 9000 training steps.

\paragraph{Posterior sampling}
% To solve inverse problems like denoising and deblurring, we leverage diffusion posterior sampling (DPS)~\cite{chung2022diffusion} algorithm to sample from given observation.
In addition to generating posterior samples using amortized inference, the inverse problem can also be solved by leveraging the Diffusion Posterior Sampling (DPS) algorithm~\cite{chung2022diffusion} to sample from corrupted observations using the learned diffusion model. Specifically, we modify the reverse-time SDE from Eq.~\ref{eq:diffusionSDE} into a conditional reversed process for posterior sampling:
\begin{equation}
    \quad d\mathbf{x} = \left[\mathbf{f}(\mathbf{x},t) - g(t)^2\nabla_\mathbf{x} \log p_t(\mathbf{x}\mid\mathbf{y})\right]dt + g(t)d\mathbf{\overline{w}},
    \label{eq:conditional-score-inverse}
\end{equation}
where $\mathbf{y}$ is the given observation. Using Bayes' theorem, the conditional score function can be decomposed into:
\begin{equation}
    \nabla_\mathbf{x} \log p_t(\mathbf{x}|\mathbf{y})= \nabla_\mathbf{x} \log p_t(\mathbf{y}\mid\mathbf{x}) +\nabla_\mathbf{x} \log p_t(\mathbf{x}).
    \label{eq:score-obs}
\end{equation}
Here $\nabla_\mathbf{x} \log p_t(\mathbf{x})$ can be replaced by the learned score function $\mathbf{s}_{\theta}(\mathbf{x},t)$, while for $\nabla_\mathbf{x} \log p_t(\mathbf{y}\mid\mathbf{x})$ we adopt the approximation proposed by DPS~\cite{chung2022diffusion}:
\begin{equation}
    p_{t}\left(\mathbf{y}\mid\mathbf{x}\right) \simeq p\left(\mathbf{y}\mid\hat{\mathbf{x}}_{0}(\mathbf{x})\right), \quad 
    \text{where} \quad  \hat{\mathbf{x}}_{0}(\mathbf{x}_{t}) := \mathbb{E}\left[\mathbf{x}_{0} \mid \mathbf{x}_{t}\right],
    \label{eq:innerexpect}
\end{equation}
Substituting Eq.\ref{eq:innerexpect} and Eq.\ref{eq:score-obs} into Eq.~\ref{eq:conditional-score-inverse}, we obtain a conditional reverse-time SDE that can reconstruct images from corrupted observations.
% Different from generative models like normalizing flows, diffusion models have state-of-the-art sampling techniques and do not require training in the sampling process, thus having faster sampling speed and better reconstruction performances.

\begin{figure}[tbp]
    \centering
    \includegraphics[scale=0.48]{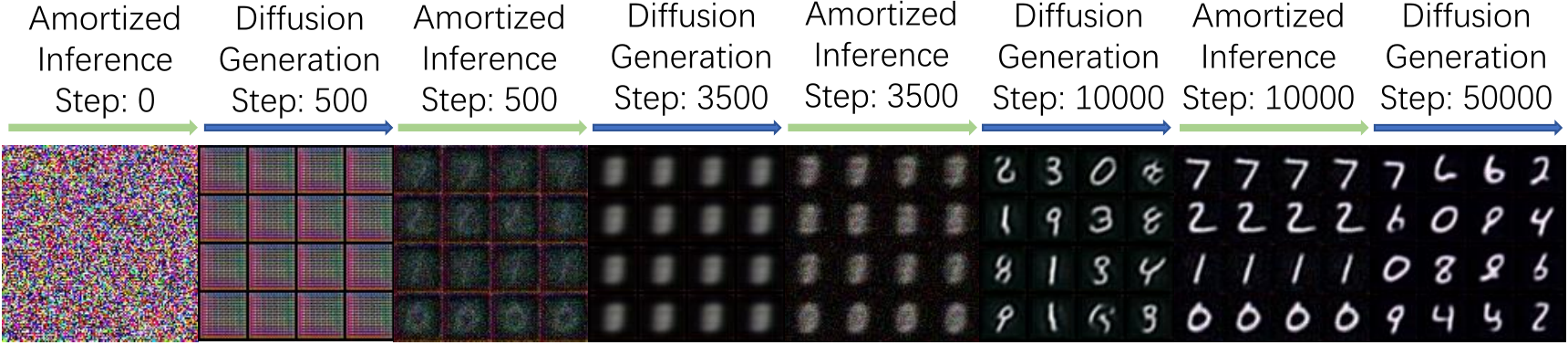}
    \caption{\textbf{Training procedure of the conditional flow model and the diffusion model.} We alternately report the amortized inference results from the flow model and the generative images from the diffusion model during training. The diffusion model initially captures low-frequency signals, guiding the amortized inference model. As the amortized inference improves, it produces better-quality images, further enhancing the diffusion model's training. Eventually, both models converge to produce clean images.}
    \label{fig:jointtraining}
\end{figure}

\section{Experiments}
In this section, we first demonstrate our method on image denoising and deblurring tasks using various datasets, including MNIST, CIFAR-10, and fluorescent microscopic images of tubulins. After that, we apply the models learned by our methods to solving inverse problems.
Further details on neural network architectures, training settings, and additional reconstruction and generation samples are provided in the appendix.

\subsection{Experimental setting}
\label{subsec:expsetting}
\paragraph{Datasets}
Our experiments are conducted on three sets of corrupted observations. First, we performed a toy experiment on denoising MNIST images corrupted by additive Gaussian noise with $\sigma=0.3$. Next, we tested our method on a deblurring task using dog images from CIFAR-10, where all images were blurred by a Gaussian kernel of size 3$\times$3 and a standard deviation of 1.5 pixels. Finally, for more realistic, higher-resolution images, we attempted to restore and learn a clean distribution from noisy microscopic images of tubulins, assuming they are corrupted by additive Gaussian noise with $\sigma=0.2$.

% We demonstrate the utilities of our methods on corrupted observations. We first ran a toy experiment of noisy MNIST images with the noise level $\sigma=0.3$ added to the original dataset. Furthermore, for complex data distributions, we select the dog category of CIFAR10: We solve image deblur of CIFAR10 with kernel size equals $(3,3)$, and intensity equals 1.5. As for high-resolution images, we use noisy biological data with the noise level $\sigma=0.2$ the resolution of the images is $64 \times 64$.

\paragraph{Evaluation metrics}
We use the Frechet Inception Distance (FID)~\cite{parmar2022aliased} to assess the generative ability of our learned diffusion model by comparing 5000 image samples generated from it to reserved test data from the underlying true distribution. For posterior sampling results, including those from either the amortized inference network or the conditional diffusion process using the learned generative model, we compute Peak Signal-to-Noise Ratio (PSNR), Structural Similarity (SSIM), and Learned Perceptual Image Patch Similarity (LPIPS) to evaluate the image reconstruction quality.
% For every sampling technique, we randomly draw samples, compute their respective metrics, and subsequently report those metrics based on the average of all the samples.

% main model's performance by sampling 5000 images and compare to the original dataset.As for posterior sampling, we computed Peak Signal to Noise Ratio(PSNR), and structural similarity(SSIM) and Learned Perceptual Image Patch Similarity(LPIPS) using the original images and our model's outputs.
% \begin{figure}[htbp]
%     \centering
%    \includegraphics[scale=0.72]{Styles/figs/prior.pdf}
%     \caption{\textbf{Image samples from diffusion models learned from corrupted observations.} The three rows show results from models trained on different datasets: noisy MNIST handwritten digits, blurred CIFAR-10 dog images, and noisy fluorescent microscope images. The learned diffusion models generate samples similar to the ground-truth images, significantly outperforming the baseline, AmbientFlow. Notably, when directly training the diffusion model using blurred images (2nd row (b)), we achieve samples with low FID scores. This is because FID mainly measures the similarity of smoothed features among image sets. However, our method (2nd row (d)) produces more reasonable and sharper dog images, despite the FID score not being superior.}
%     \label{PIC:fid}
% \end{figure}

\begin{figure*}
	% \vspace*{-0.5cm}
	\centering
	\setlength{\tabcolsep}{1pt}
	\setlength{\fboxrule}{1pt}
	%\vspace*{1.5cm}
	\begin{tabular}{c}
		\begin{tabular}{cccc}
			% & 
			% \tiny{\makecell[c]{Noisy\\Observation}} & 
			% \tiny{\makecell[c]{SURE-\\Score~\cite{aali2023solving}}} & 
			% \tiny{\makecell[c]{Ambient\\Diffusion~\cite{daras2023ambient}}}
			% \\ 
			% \begin{turn}{90} \!\!\! \!\!\! \!\!\! \!\!\! \!\!\! \!\!\!\small{ImageNet} \end{turn} & 
			\multicolumn{1}{c}{
				\begin{overpic}[width=0.245\linewidth]{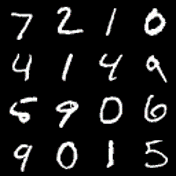}
				\end{overpic}
			}  &
                \multicolumn{1}{c}{
				\begin{overpic}[width=0.245\linewidth]{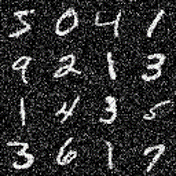}
				\end{overpic}
			}  &
                \multicolumn{1}{c}{
				\begin{overpic}[width=0.245\linewidth]{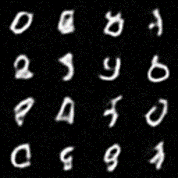}
				\end{overpic}
			}  &
                \multicolumn{1}{c}{
				\begin{overpic}[width=0.245\linewidth]{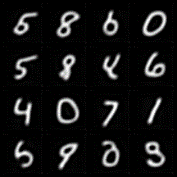}
				\end{overpic}
			}
                \\[-0.5ex]
                \multicolumn{1}{c}{(a) Ground Truth} &
                {\makecell[c]{(b) Observations,\\FID=204}}&
                % \multicolumn{1}{c}{(b) Observations, FID=204} &
                {\makecell[c]{(c) AmbientFlow~\cite{kelkar2023ambientflow},\\FID=147}}&
                {\makecell[c]{(d) Ours,\\FID=149}}
                % \multicolumn{1}{c}{(c) AmbientFlow, FID=147} &
                % \multicolumn{1}{c}{(d) Ours, FID=149}
                \\
                \\[-2ex]
                \multicolumn{1}{c}{
				\begin{overpic}[width=0.245\linewidth]{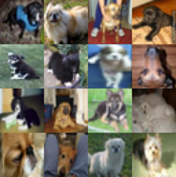}
				\end{overpic}
			}  &
                \multicolumn{1}{c}{
				\begin{overpic}[width=0.245\linewidth]{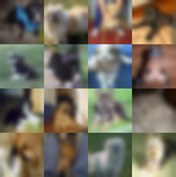}
				\end{overpic}
			}  &
                \multicolumn{1}{c}{
				\begin{overpic}[width=0.245\linewidth]{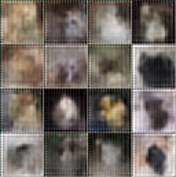}
				\end{overpic}
			}  &
                \multicolumn{1}{c}{
				\begin{overpic}[width=0.245\linewidth]{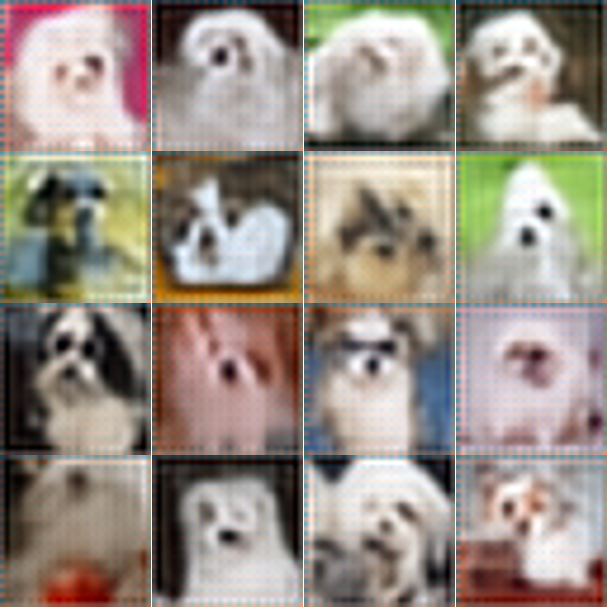}
				\end{overpic}
			}
                \\[-0.5ex]
                \multicolumn{1}{c}{(a) Ground Truth} &
                {\makecell[c]{(b) Observations,\\FID=109}}&
                % \multicolumn{1}{c}{(b) Observations, FID=109} &
                {\makecell[c]{(c) AmbientFlow~\cite{kelkar2023ambientflow},\\FID=272}}&
                {\makecell[c]{(d) Ours,\\FID=209}}
                % \multicolumn{1}{c}{(c) AmbientFlow, FID=272} &
                % \multicolumn{1}{c}{(d) Ours, FID=209}
                \\
                \\[-2.0ex]
                \multicolumn{1}{c}{
				\begin{overpic}[width=0.245\linewidth]{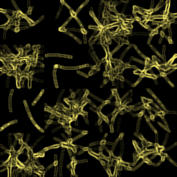}
				\end{overpic}
			}  &
                \multicolumn{1}{c}{
				\begin{overpic}[width=0.245\linewidth]{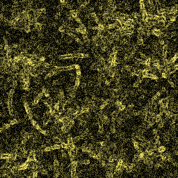}
				\end{overpic}
			}  &
                \multicolumn{1}{c}{
				\begin{overpic}[width=0.245\linewidth]{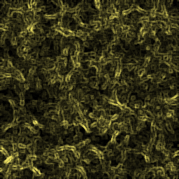}
				\end{overpic}
			}  &
                \multicolumn{1}{c}{
				\begin{overpic}[width=0.245\linewidth]{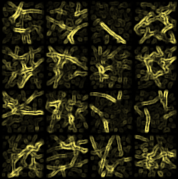}
				\end{overpic}
			}
                \\[-0.5ex]
                \multicolumn{1}{c}{(a) Ground Truth} &
                {\makecell[c]{(b) Observations,\\FID=298}}&
                % \multicolumn{1}{c}{(b) Observations, FID=298} &
                {\makecell[c]{(c) AmbientFlow~\cite{kelkar2023ambientflow},\\FID=295}}&
                {\makecell[c]{(d) Ours,\\FID=210}}
                % \multicolumn{1}{c}{(c) AmbientFlow, FID=295} &
                % \multicolumn{1}{c}{(d) Ours, FID=210}
		\end{tabular}
	\end{tabular}
        \caption{\textbf{Image samples from diffusion models learned from corrupted observations.} The three rows show results from models trained on different datasets: noisy MNIST handwritten digits, blurred CIFAR-10 dog images, and noisy fluorescent microscope images. The learned diffusion models generate samples similar to the ground-truth images, significantly outperforming the baseline, AmbientFlow. Notably, when directly training the diffusion model using blurred images (2nd row (b)), we achieve samples with low FID scores. This is because FID mainly measures the similarity of smoothed features among image sets. However, our method (2nd row (d)) produces more reasonable and sharper dog images, despite the FID score not being superior.}
    \label{PIC:fid}
\end{figure*}

\paragraph{Baselines}
We compare our methods with three baselines that operate under the same conditions, where no clean signals are available. AmbientFlow~\cite{kelkar2023ambientflow} employs a similar amortized inference approach but uses a normalizing flow model~\cite{kingma2018glow}, instead of a diffusion model, to learn the unconditional clean distribution. AmbientDiffusion~\cite{daras2023ambient} learns a clean score-based prior by further corrupting the input data, rather than restoring the clean images first. SURE-Score~\cite{aali2023solving} combines the SURE loss~\cite{stein1981estimation} to implicitly regularize the weights of the learned diffusion model, enabling direct training of clean diffusion models using corrupted observations. We carefully tune the hyperparameters of all the baselines and report the best results. More information on these neural networks' architectures and hyperparameters can be found in Appendix~\ref{app:architecture}.

\subsection{Results}
% \begin{figure}[htbp]
%     \centering
%     \includegraphics[scale=0.7]{Styles/figs/prior.pdf}
%     \caption{Both visual results and numerical results of the main model from our method and AmbientFlow. The trianing datasets we select are: MNIST corrupted with Gaussian noise with $\sigma=0.3$, CIFAR10 (dog category) blurred by  Gaussian blur kernel $3 \times 3$ kernel size and $\sigma=0.5 $, and Synthetic biological data of tubulins with $\sigma=0.3$ Gaussian noise added. Our method successfully modelled the clean data distribution on all three tasks.}
%     \label{PIC:fid}
% \end{figure}

% \begin{figure}[htbp]
%     \centering
%     \includegraphics[scale=0.95]{Styles/figs/post-model.pdf}
%     \caption{\textbf{Amortized inference results on CIFAR-10 deblurring and microscopy imaging tasks.} Our method achieves superior performance compared to AmbientFlow, because of the diffusion model's stronger generative modeling capabilities over the flow model employed by AmbientFlow.
%     % Output samples from the conditional invertible network. Quantitative results show that the posterior flow model successfully improves the image quality. Our method produces better outputs compared to AmbientFlow. This suggests that both AmbientFlow and our method's conditional invertible models are able to generate clean images. However, our method outperforms AmbientFlow due to the strong modeling capacity of the DM we choose.
%     }
%     \label{fig:post-model-output}
% \end{figure}

\begin{figure*}
	% \vspace*{-0.5cm}
	\centering
	\setlength{\tabcolsep}{1pt}
	\setlength{\fboxrule}{1pt}
	%\vspace*{1.5cm}
	\begin{tabular}{c}
		\begin{tabular}{cccc|cccc}
			% & 
			\small{\makecell[c]{Observation}} & 
                \small{\makecell[c]{Ambient\\Flow~\cite{kelkar2023ambientflow}}} &
			\small{\makecell[c]{Ours}} & 
			\small{\makecell[c]{Ground \\ Truth}}&
                \small{\makecell[c]{Observation}} & 
                \small{\makecell[c]{Ambient\\Flow~\cite{kelkar2023ambientflow}}} &
			\small{\makecell[c]{Ours}} & 
			\small{\makecell[c]{Ground \\ Truth}}\cr
			% \begin{turn}{90} \!\!\! \!\!\! \!\!\! \!\!\! \!\!\! \!\!\!\small{ImageNet} \end{turn} & 
			\multicolumn{1}{c}{
				\begin{overpic}[width=0.120\linewidth]{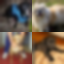}
				\end{overpic}
			}  &
                \multicolumn{1}{c}{
				\begin{overpic}[width=0.120\linewidth]{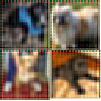}
				\end{overpic}
			}  &
                \multicolumn{1}{c}{
				\begin{overpic}[width=0.120\linewidth]{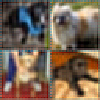}
				\end{overpic}
			}  &
			\multicolumn{1}{c|}{
				\begin{overpic}[width=0.120\linewidth]{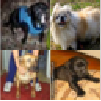}
				\end{overpic}
			}  &
			\multicolumn{1}{c}{
				\begin{overpic}[width=0.120\linewidth]{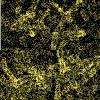}
				\end{overpic}
			}  &
			\multicolumn{1}{c}{
				\begin{overpic}[width=0.120\linewidth]{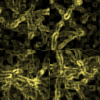}
				\end{overpic}
			}  &
			\multicolumn{1}{c}{
				\begin{overpic}[width=0.120\linewidth]{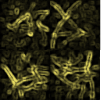}
				\end{overpic}
			}  &
                \multicolumn{1}{c}{
				\begin{overpic}[width=0.120\linewidth]{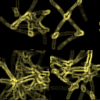}
				\end{overpic}
			} 
			  \\
                \multicolumn{4}{c|}{(a) Amortized Inference Results of CIFAR-10} &
                \multicolumn{4}{c}{(b) Amortized Inference Results of Tubulins} 
		\end{tabular}
	\end{tabular}
	\caption{\textbf{Amortized inference results on CIFAR-10 deblurring and microscopy imaging tasks.} Our method achieves superior performance compared to AmbientFlow, because of the diffusion model's stronger generative modeling capabilities over the flow model employed by AmbientFlow.}
\label{fig:post-model-output}
% \vspace{-0.1in}
\end{figure*}

\begin{figure*}
	% \vspace*{-0.5cm}
	\centering
	\setlength{\tabcolsep}{1pt}
	\setlength{\fboxrule}{1pt}
	%\vspace*{1.5cm}
	\begin{tabular}{c}
		\begin{tabular}{cccccc|cccccc}
			% & 
			\tiny{\makecell[c]{Observation}} & 
                \tiny{\makecell[c]{Ambient\\Diffusion~\cite{daras2023ambient}}} &
			\tiny{\makecell[c]{SURE-\\Score~\cite{aali2023solving}}} & 
			\tiny{\makecell[c]{Ambient\\Flow~\cite{kelkar2023ambientflow}}}&
			\tiny{\makecell[c]{Ours}}&
			\tiny{\makecell[c]{Ground\\Truth}} &
                \tiny{\makecell[c]{Observation}} & 
                \tiny{\makecell[c]{Ambient\\Diffusion~\cite{daras2023ambient}}} &
			\tiny{\makecell[c]{SURE-\\Score~\cite{aali2023solving}}} & 
			\tiny{\makecell[c]{Ambient\\Flow~\cite{kelkar2023ambientflow}}}&
			\tiny{\makecell[c]{Ours}}&
			\tiny{\makecell[c]{Ground\\Truth}} 
			\\ 
			% \begin{turn}{90} \!\!\! \!\!\! \!\!\! \!\!\! \!\!\! \!\!\!\small{ImageNet} \end{turn} & 
			\multicolumn{1}{c}{
				\begin{overpic}[width=0.078\linewidth]{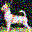}
				\end{overpic}
			}  &
                \multicolumn{1}{c}{
				\begin{overpic}[width=0.078\linewidth]{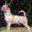}
				\end{overpic}
			}  &
                \multicolumn{1}{c}{
				\begin{overpic}[width=0.078\linewidth]{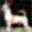}
				\end{overpic}
			}  &
			\multicolumn{1}{c}{
				\begin{overpic}[width=0.078\linewidth]{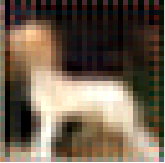}
				\end{overpic}
			}  &
			\multicolumn{1}{c}{
				\begin{overpic}[width=0.078\linewidth]{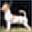}
				\end{overpic}
			}  &
			\multicolumn{1}{c|}{
				\begin{overpic}[width=0.078\linewidth]{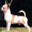}
				\end{overpic}
			}  &
			\multicolumn{1}{c}{
				\begin{overpic}[width=0.078\linewidth]{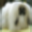}
				\end{overpic}
			}  &
			\multicolumn{1}{c}{
				\begin{overpic}[width=0.078\linewidth]{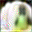}
				\end{overpic}
			}  &
			\multicolumn{1}{c}{
				\begin{overpic}[width=0.078\linewidth]{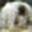}
				\end{overpic}
			}  &
                \multicolumn{1}{c}{
				\begin{overpic}[width=0.078\linewidth]{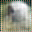}
				\end{overpic}
			}  &
                \multicolumn{1}{c}{
				\begin{overpic}[width=0.078\linewidth]{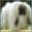}
				\end{overpic}
			}  &
			\multicolumn{1}{c}{
				\begin{overpic}[width=0.078\linewidth]{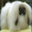}
				\end{overpic}
			} 
			\\
			  \multicolumn{1}{c}{
			  	\begin{overpic}[width=0.078\linewidth]{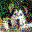}
			  	\end{overpic}
			  }  &
                \multicolumn{1}{c}{
				\begin{overpic}[width=0.078\linewidth]{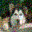}
				\end{overpic}
			}  &
                \multicolumn{1}{c}{
				\begin{overpic}[width=0.078\linewidth]{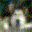}
				\end{overpic}
			}  &
			  \multicolumn{1}{c}{
			  	\begin{overpic}[width=0.078\linewidth]{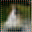}
			  	\end{overpic}
			  }  &
			  \multicolumn{1}{c}{
			  	\begin{overpic}[width=0.078\linewidth]{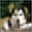}
			  	\end{overpic}
			  }  &
			  \multicolumn{1}{c|}{
			  	\begin{overpic}[width=0.078\linewidth]{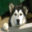}
			  	\end{overpic}
			  }  &
			  \multicolumn{1}{c}{
			  	\begin{overpic}[width=0.078\linewidth]{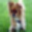}
			  	\end{overpic}
			  }  &
			  \multicolumn{1}{c}{
			  	\begin{overpic}[width=0.078\linewidth]{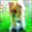}
			  	\end{overpic}
			  }  &
			  \multicolumn{1}{c}{
			  	\begin{overpic}[width=0.078\linewidth]{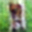}
			  	\end{overpic}
			  }  &
                \multicolumn{1}{c}{
			  	\begin{overpic}[width=0.078\linewidth]{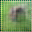}
			  	\end{overpic}
			  }  &
                \multicolumn{1}{c}{
			  	\begin{overpic}[width=0.078\linewidth]{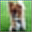}
			  	\end{overpic}
			  }  &
			  \multicolumn{1}{c}{
			  	\begin{overpic}[width=0.078\linewidth]{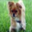}
			  	\end{overpic}
			  } 
			  \\
                \multicolumn{6}{c|}{(a) CIFAR-10, Denoising} &
                \multicolumn{6}{c}{(b) CIFAR-10, Deblurring} 
                \\ \\
                % & 
				\tiny{\makecell[c]{Observation}} & 
                \tiny{\makecell[c]{Ambient\\Diffusion~\cite{daras2023ambient}}} &
			\tiny{\makecell[c]{SURE-\\Score~\cite{aali2023solving}}} & 
			\tiny{\makecell[c]{Ambient\\Flow~\cite{kelkar2023ambientflow}}}&
			\tiny{\makecell[c]{Ours}}&
			\tiny{\makecell[c]{Ground\\Truth}} &
                \tiny{\makecell[c]{Observation}} & 
                \tiny{\makecell[c]{Ambient\\Diffusion~\cite{daras2023ambient}}} &
			\tiny{\makecell[c]{SURE-\\Score~\cite{aali2023solving}}} & 
			\tiny{\makecell[c]{Ambient\\Flow~\cite{kelkar2023ambientflow}}}&
			\tiny{\makecell[c]{Ours}}&
			\tiny{\makecell[c]{Ground\\Truth}} 
			\\
			% \begin{turn}{90} \!\!\! \!\!\! \!\!\! \!\!\! \!\!\! \!\!\!\small{ImageNet} \end{turn} & 
			\multicolumn{1}{c}{
				\begin{overpic}[width=0.078\linewidth]{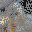}
				\end{overpic}
			}  &
                \multicolumn{1}{c}{
				\begin{overpic}[width=0.078\linewidth]{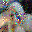}
				\end{overpic}
			}  &
                \multicolumn{1}{c}{
				\begin{overpic}[width=0.078\linewidth]{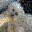}
				\end{overpic}
			}  &
			\multicolumn{1}{c}{
				\begin{overpic}[width=0.078\linewidth]{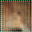}
				\end{overpic}
			}  &
			\multicolumn{1}{c}{
				\begin{overpic}[width=0.078\linewidth]{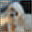}
				\end{overpic}
			}  &
			\multicolumn{1}{c|}{
				\begin{overpic}[width=0.078\linewidth]{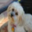}
				\end{overpic}
			}  &
			\multicolumn{1}{c}{
				\begin{overpic}[width=0.078\linewidth]{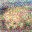}
				\end{overpic}
			}  &
			\multicolumn{1}{c}{
				\begin{overpic}[width=0.078\linewidth]{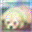}
				\end{overpic}
			}  &
			\multicolumn{1}{c}{
				\begin{overpic}[width=0.078\linewidth]{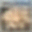}
				\end{overpic}
			}  &
                \multicolumn{1}{c}{
				\begin{overpic}[width=0.078\linewidth]{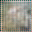}
				\end{overpic}
			}  &
                \multicolumn{1}{c}{
				\begin{overpic}[width=0.078\linewidth]{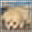}
				\end{overpic}
			}  &
			\multicolumn{1}{c}{
				\begin{overpic}[width=0.078\linewidth]{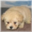}
				\end{overpic}
			} 
			\\
			  \multicolumn{1}{c}{
			  	\begin{overpic}[width=0.078\linewidth]{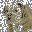}
			  	\end{overpic}
			  }  &
                \multicolumn{1}{c}{
				\begin{overpic}[width=0.078\linewidth]{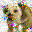}
				\end{overpic}
			}  &
                \multicolumn{1}{c}{
				\begin{overpic}[width=0.078\linewidth]{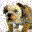}
				\end{overpic}
			}  &
			  \multicolumn{1}{c}{
			  	\begin{overpic}[width=0.078\linewidth]{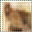}
			  	\end{overpic}
			  }  &
			  \multicolumn{1}{c}{
			  	\begin{overpic}[width=0.078\linewidth]{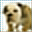}
			  	\end{overpic}
			  }  &
			  \multicolumn{1}{c|}{
			  	\begin{overpic}[width=0.078\linewidth]{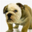}
			  	\end{overpic}
			  }  &
			  \multicolumn{1}{c}{
			  	\begin{overpic}[width=0.078\linewidth]{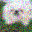}
			  	\end{overpic}
			  }  &
			  \multicolumn{1}{c}{
			  	\begin{overpic}[width=0.078\linewidth]{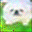}
			  	\end{overpic}
			  }  &
			  \multicolumn{1}{c}{
			  	\begin{overpic}[width=0.078\linewidth]{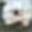}
			  	\end{overpic}
			  }  &
                \multicolumn{1}{c}{
			  	\begin{overpic}[width=0.078\linewidth]{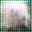}
			  	\end{overpic}
			  }  &
                \multicolumn{1}{c}{
			  	\begin{overpic}[width=0.078\linewidth]{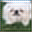}
			  	\end{overpic}
			  }  &
			  \multicolumn{1}{c}{
			  	\begin{overpic}[width=0.078\linewidth]{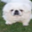}
			  	\end{overpic}
			  } 
			  \\
			 
                \multicolumn{6}{c|}{(c) CIFAR-10, Inpainting} &
                \multicolumn{6}{c}{(d) CIFAR-10, Deblurring+Denoising} 
		\end{tabular}
	\end{tabular}
	\caption{\textbf{Posterior samples from the generative model trained on blurred CIFAR-10 images.} On four downstream tasks - denoising, deblurring, inpainting, and combined denoising and deblurring - our method surpasses the performance of baseline approaches including AmbientDiffusion, SURE-Score, and AmbientFlow.}
\label{fig:cifar-posterior}
\vspace{-0.1in}
\end{figure*}

\paragraph{Clean distributions learned from corrupted observations}
Fig \ref{PIC:fid} compares the FID scores of image samples from the baseline method, AmbientFlow, and our method, FlowDiff. As shown in Fig.\ref{PIC:fid}, our method significantly outperforms AmbientFlow across all three tasks, learning high-quality, complex, clean image distributions from blurred or noisy observations. During training, as illustrated in Fig.~\ref{fig:jointtraining}, we observed that the diffusion model captures low-frequency signals first, providing guidance for the amortized inference model. As the amortized inference improves, it produces better-quality images, which in turn enhances the training of the diffusion model. By alternatively updating the weights of these models, the diffusion model eventually learns the distribution of the clean data. The model reset technique described in Sec.~\ref{sec:method-details} is used in our training process.

% can learn clean image priors from blurred or noisy observations. What's more, the FID score of Flowdiff competes or outperforms AmbientFlow on all three tasks, which shows that our framework is capable of modeling complex data distributions and high-resolution images. During the training process, we observed that the diffusion model will capture the low-frequency signals first, thus provides guidance for the posterior model. As the posterior model improves, it will produce better-quality images to the diffusion model. By updating the models' weights in turn, the diffusion model will learn the distribution of the clean data. However, for complex data distributions like CIFAR10 and the biological images, the performance of the model will not keep improving as the training goes on. To tackle this issue, we will freeze the weights of the posterior model and train the diffusion model from scratch.
\begin{table}[htbp]
  \centering
  \caption{\textbf{Amortized inference results for three different tasks.} In all cases, our method produces a better flow model for amortized inference than AmbientFlow, despite using the same flow architectures. This indicates that the diffusion model in our framework provides a superior prior compared to the flow prior in AmbientFlow. The optimal results are highlighted in \textbf{bold}.
  % Quantitative analysis of the outputs generated by the posterior models in both AmbientFlow and our framework demonstrates that both models enhance image quality. However, the posterior model in our framework exhibits superior performance compared to AmbientFlow, which suggest that with a stronger DM prior, the performance of the conditional invertible network will also be improved through amortized inference.
  }
  \setlength{\tabcolsep}{0.6mm}{
    \begin{tabular}{cccccccccc}
    \toprule
    \multirow{2}[4]{*}{\textbf{Tasks}} & \multicolumn{3}{c}{\textbf{Input}} & \multicolumn{3}{c}{\textbf{AmbientFlow}} & \multicolumn{3}{c}{\textbf{Ours}} \\
\cmidrule{2-10}          & PSNR$\uparrow $  & SSIM$\uparrow $ & LPIPS$\downarrow $   & PSNR$\uparrow $  & SSIM$\uparrow $ & LPIPS$\downarrow $   & PSNR$\uparrow $  & SSIM$\uparrow $ & LPIPS$\downarrow $ \\
    \midrule
    MNIST Denoising  & 13.57 & 0.210 & 0.591 & \textbf{21.18} & 0.394 & 0.177 & 20.73 & \textbf{0.399} & \textbf{0.160} \\
   CIFAR-10 Deblurring & 20.91 & 0.582 & 0.182 & 20.38 & 0.704 & 0.199 & \textbf{21.97} & \textbf{0.787} & \textbf{0.135} \\
  Microscopy Imaging & 14.32 & 0.106 & 0.572 & 16.76 & 0.220 & 0.467 & \textbf{18.87} & \textbf{0.263} & \textbf{0.397} \\
    \bottomrule
    \end{tabular}}
  \label{tab:post-model-results}%
\end{table}%

\paragraph{Amortized inference}

Fig.~\ref{fig:post-model-output} and Table~\ref{tab:post-model-results} present the amortized inference results, i.e., the posterior samples drawn from the conditional normalizing flow. Our method produces better reconstructed images compared to AmbientFlow. Furthermore, results show that the amortized inference models trained by our method achieve performance comparable to those trained with a clean diffusion prior, again demonstrating that our method successfully captures the underlying clean data distribution. More details can be found in Appendix~\ref{app:clean}.

% output samples from the conditional invertible network, which is used to estimate the posterior distribution in our amortized inference framework. Both numerical and visual results shows that the posterior model trained under our framework generates better quality images compared to Ambientflow. Further results show that the model is compatible with the posterior flow model trained with a pre-trained clean diffusion prior. The result suggests that our framework successfully captures the underlying clean data distribution. 

% Table generated by Excel2LaTeX from sheet 'Sheet2'
\begin{table}[htbp]
  \centering
  \setlength{\tabcolsep}{1.0mm}
  {\caption{\textbf{Performance on downstream posterior sampling tasks using the CIFAR-10 dog images.} The optimal results are highlighted in \textbf{bold}, and the second-best results are \underline{underlined}. All metrics are computed using 128 posterior samples.}
    \begin{tabular}{ccccccc}
    \toprule
    \textbf{Tasks} & \textbf{Metrics} & \textbf{Input} & \textbf{\makecell[c]{Ambient \\ Diffusion}} & \textbf{\makecell[c]{SURE- \\ Score}} & \textbf{\makecell[c]{Ambient \\ Flow}} & \textbf{Ours} \\
    \midrule
    \multirow{3}[2]{*}{Denoising} & PSNR $\uparrow$  &     14.79  &   \underline{21.37}    &  19.04     &    14.95   & \textbf{21.70} \\
          & SSIM $\uparrow$ &   0.444    &  \underline{0.743}     &  0.636     &   0.356    & \textbf{0.782}  \\
          & LPIPS $\downarrow$ &  0.107     &   \textbf{0.033}     &    0.111    &   0.142    & \underline{0.040} \\
    \midrule
    \multirow{3}[2]{*}{\makecell[c]{Deblurring}} & PSNR$\uparrow$  &   22.34    &    16.14   &    \underline{23.52}   &  15.03     & \textbf{23.91}  \\
          & SSIM $\uparrow$ & 0.724      &  0.761     &  \underline{0.829}      &  0.305     & \textbf{0.880} \\
          & LPIPS$\downarrow$ &   0.228    &  \underline{0.074}     &  0.078     &  0.163     & \textbf{0.031} \\
    \midrule
    \multirow{3}[2]{*}{\makecell[c]{Deblurring + Denoising} }& PSNR$\uparrow$  &   19.34    &  16.23     &  \underline{21.28}     &   15.11    &  \textbf{22.55}\\
          & SSIM $\uparrow$ &   0.609    &     \underline{0.767}  &  0.666     &  0.339     &  \textbf{0.816}\\
          & LPIPS$\downarrow$ &   0.044    &   \underline{0.062}    &     0.122  &   0.137    & \textbf{0.037}  \\
    \midrule
    \multirow{3}[2]{*}{Inpainting} & PSNR $\uparrow$ &   13.49     &  20.57     &   \underline{21.60}    &      13.20  & \textbf{22.46} \\
          & SSIM $\uparrow$ &   0.404    &  0.639     &  \underline{0.732}     &  0.217     & \textbf{0.836}  \\
          & LPIPS$\downarrow$ &  0.295     &  \underline{0.038}     &  0.049     &   0.179    & \textbf{0.034} \\
    \bottomrule
    \end{tabular}%
  \label{tab:cifartable}}
\end{table}%

\paragraph{Posterior sampling with learned clean prior}
We leverage the learned clean priors to solve various downstream computational imaging inverse problems, including inpainting, denoising, deblurring, and a combination of denoising and deblurring. Table~\ref{tab:cifartable} and Fig.~\ref{fig:cifar-posterior} present comprehensive experiments on CIFAR-10 across all four tasks. Assuming the clean distributions are trained on blurred images as explained in Sec.~\ref{subsec:expsetting}, our methods significantly outperform all the baselines, including AmbientDiffusion, SURE-Score, and AmbientFlow, across all tasks.

\section{Conclusion and limitation}
In this work, we present FlowDiff, a framework that integrates amortized inference with state-of-the-art diffusion models to learn clean signal distributions directly from corrupted observations. Through amortized inference, our framework incorporates an additional normalizing flow~\cite{kingma2018glow} that generates clean images from corrupted observations. The normalizing flow is trained jointly with the DM in a variational inference framework: the normalizing flow generates clean images for training the DM, while the DM imposes an image prior to guide reasonable estimations by the normalizing flow. After training, our method provides both a diffusion prior that models a complex, high-quality clean image distribution and a normalizing flow-based amortized inference network that directly generates posterior samples from corrupted observations. We demonstrate our method through extensive experiments on various datasets and multiple computational imaging tasks. We also apply the models our method learned to solving inverse problems including denoising, deblurring, and inpainting.

However, the different learning speeds of diffusion models and normalizing flows make the joint training of the two networks sometimes unstable. Besides, normalizing flows often fail to model complex data distributions due to their limited model capacity. In the future, we plan to explore better optimization frameworks, such as alternating optimization methods like expectation-maximization, to achieve more stable training. 
% We can also alternate the flow model into a better generative model that can calculate explicit probability. 
% We plan to apply this framework to real scientific and biomedical imaging problems, such as super-resolution microscopy and computed tomography.

% In addition, the surrogate bound in Eq.~\ref{} is only an approximation of the image probability. the probability of images is unable to model noises as correctly as AmbientFlow. Further improvements can be made to adapt a better main model that can estimate exact prior, or find a more stable training scheme.

% Thanks to the strong power of diffusion models, our method can learn a complex, high-quality and clean image distribution

% a jointly optimizing framework leveraging state-of-the-art diffusion models to learn clean signals directly from noisy data. Due to the strong capacity of DDPM, our method can model complex distributions and high-resolution images. However, the different learning speeds of diffusion models and normalization flows make it hard to optimize. What's more, the surrogate bound we use to calculate the probability of images is unable to model noises as correctly as AmbientFlow. Further improvements can be made to adapt a better main model that can estimate exact prior, or find a more stable training scheme.

% \newpage

%%%%%%%%%%%%%%%%%%%%%%%%%%%%%%%%%%%%%%%%%%%%%%%%%%%%%%%%%%%%

%%%%%%%%%%%%%%%%%%%%%%%%%%%%%%%%%%%%%%%%%%%%%%%%%%%%%%%%%%%%

\newpage
\appendix

\section{Neural network architectures and hyper-parameters}
\label{app:architecture}

We conducted all our experiments using the same neural network architectures, but slightly adjusting the number of parameters based on the complexity of the data distribution. The specific numbers are shown in Table \ref{TAB:parameter}. For our amortized inference model, we use the same conditional invertible network as AmbientFlow~\cite{kelkar2023ambientflow} and employ DDPM~\cite{ho2020denoising} for learning the clean distribution. To balance the learning speeds between the flow model and the diffusion model, we set the learning rate of the diffusion model to be smaller. Specifically, for the MNIST and CIFAR-10 experiments, the flow network's learning rate is $1e-3$ and the diffusion model's learning rate is $1e-4$. For the microscopic experiment, we train the normalizing flow with a learning rate of $2e-5$ and the diffusion model with a learning rate of $1e-5$. We use Adam~\cite{kingma2014adam} as the optimizer for our models. All the experiments are conducted on an NVIDIA A800 GPU workstation.

\begin{table}[!ht]
    \caption{The number of model parameters for each experiment. We adjusted the size of our models based on the complexity of data distributions as well as the resolution of input images.}
    \centering
    \begin{tabular}{cccc}
    \hline
        ~ & MNIST & CIFAR-10 & \makecell[c]{Microscopic images\\ of tubulins} \\ \hline
        Flow & 8M & 11.5M & 23.9M \\ \hline
        Diffusion & 6M & 35.7M & 17.2M \\ \hline
    \end{tabular}
    \label{TAB:parameter}
\end{table}

% \begin{figure}[hbt!]
%     \centering
%     \includegraphics[scale=0.80]{Styles/figs/bio-post.pdf}
%     \caption{\textbf{Posterior samples from the generative model trained on noisy microscopic images.} On the downstream image denoising task, our approach outperforms BM3D~\cite{dabov2007image} and AmbientFlow~\cite{kelkar2023ambientflow} by effectively removing noise while preserving intricate structural details of microscopic images of tubulins.
%     % Denoising task on synthetic biological data of tubulins, noise level $\sigma=0.2$. The tubulins are highlighted by yellow fluorescence marks. The images are resized to $64 \time 64$. Due to the complexity of tubulins, our approach outperforms both BM3D and AmbientFlow.
%     }
%     \label{fig:bio-posterior}
% \end{figure}

\begin{figure*}
	% \vspace*{-0.5cm}
	\centering
	\setlength{\tabcolsep}{1pt}
	\setlength{\fboxrule}{1pt}
	%\vspace*{1.5cm}
	\begin{tabular}{c}
		\begin{tabular}{ccccc}
			% & 
			% \tiny{\makecell[c]{Noisy\\Observation}} & 
			% \tiny{\makecell[c]{SURE-\\Score~\cite{aali2023solving}}} & 
			% \tiny{\makecell[c]{Ambient\\Diffusion~\cite{daras2023ambient}}}
			% \\ 
			% \begin{turn}{90} \!\!\! \!\!\! \!\!\! \!\!\! \!\!\! \!\!\!\small{ImageNet} \end{turn} & 
			\multicolumn{1}{c}{
				\begin{overpic}[width=0.195\linewidth]{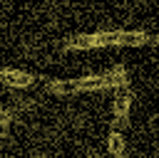}
                \put(15, 86){\textcolor{white}{PSNR: 18.16}}
				\end{overpic}
			}  &
                \multicolumn{1}{c}{
				\begin{overpic}[width=0.195\linewidth]{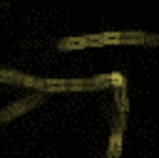}
                \put(15, 86){\textcolor{white}{PSNR: 20.77}}
				\end{overpic}
			}  &
                \multicolumn{1}{c}{
				\begin{overpic}[width=0.195\linewidth]{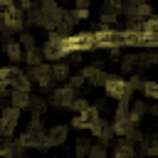}
                \put(15, 86){\textcolor{white}{PSNR: 17.74}}
				\end{overpic}
			}  &
                \multicolumn{1}{c}{
				\begin{overpic}[width=0.195\linewidth]{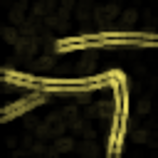}
                \put(15, 86){\textcolor{white}{PSNR: 22.87}}
				\end{overpic}
			}  &
                \multicolumn{1}{c}{
				\begin{overpic}[width=0.195\linewidth]{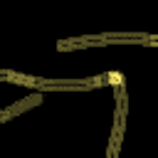}
				\end{overpic}
			}
                \\
                \multicolumn{1}{c}{
				\begin{overpic}[width=0.195\linewidth]{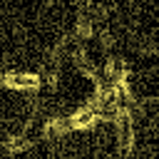}
                \put(15, 86){\textcolor{white}{PSNR: 18.00}}
				\end{overpic}
			}  &
                \multicolumn{1}{c}{
				\begin{overpic}[width=0.195\linewidth]{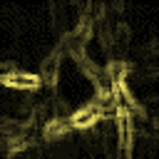}
                \put(15, 86){\textcolor{white}{PSNR: 20.38}}
				\end{overpic}
			}  &
                \multicolumn{1}{c}{
				\begin{overpic}[width=0.195\linewidth]{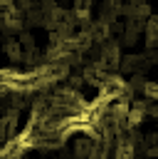}
                \put(15, 86){\textcolor{white}{PSNR: 17.33}}
				\end{overpic}
			}  &
                \multicolumn{1}{c}{
				\begin{overpic}[width=0.195\linewidth]{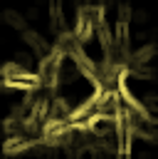}
                \put(15, 86){\textcolor{white}{PSNR: 22.99}}
				\end{overpic}
			}  &
                \multicolumn{1}{c}{
				\begin{overpic}[width=0.195\linewidth]{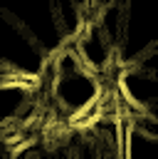}
				\end{overpic}
			}
                \\
                \multicolumn{1}{c}{
				\begin{overpic}[width=0.195\linewidth]{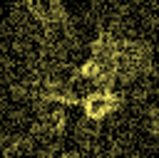}
                \put(15, 86){\textcolor{white}{PSNR: 18.12}}
				\end{overpic}
			}  &
                \multicolumn{1}{c}{
				\begin{overpic}[width=0.195\linewidth]{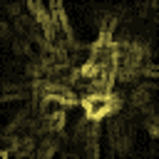}
                \put(15, 86){\textcolor{white}{PSNR: 20.28}}
				\end{overpic}
			}  &
                \multicolumn{1}{c}{
				\begin{overpic}[width=0.195\linewidth]{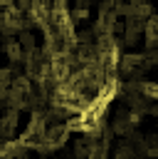}
                \put(15, 86){\textcolor{white}{PSNR: 16.87}}
				\end{overpic}
			}  &
                \multicolumn{1}{c}{
				\begin{overpic}[width=0.195\linewidth]{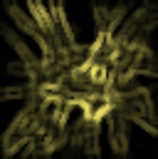}
                \put(15, 86){\textcolor{white}{PSNR: 22.63}}
				\end{overpic}
			}  &
                \multicolumn{1}{c}{
				\begin{overpic}[width=0.195\linewidth]{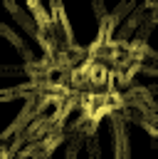}
				\end{overpic}
			}
                \\
                \multicolumn{1}{c}{
				\begin{overpic}[width=0.195\linewidth]{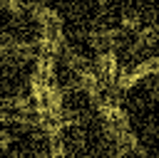}
                \put(15, 86){\textcolor{white}{PSNR: 17.94}}
				\end{overpic}
			}  &
                \multicolumn{1}{c}{
				\begin{overpic}[width=0.195\linewidth]{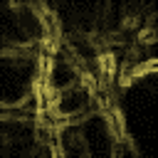}
                \put(15, 86){\textcolor{white}{PSNR: 16.01}}
				\end{overpic}
			}  &
                \multicolumn{1}{c}{
				\begin{overpic}[width=0.195\linewidth]{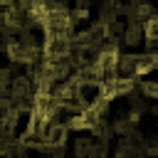}
                \put(15, 86){\textcolor{white}{PSNR: 17.10}}
				\end{overpic}
			}  &
                \multicolumn{1}{c}{
				\begin{overpic}[width=0.195\linewidth]{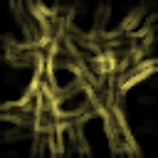}
                \put(15, 86){\textcolor{white}{PSNR: 22.56}}
				\end{overpic}
			}  &
                \multicolumn{1}{c}{
				\begin{overpic}[width=0.195\linewidth]{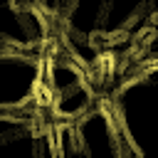}
				\end{overpic}
			}
                \\
                \multicolumn{1}{c}{Observations} &
                {\makecell[c]{BM3D~\cite{dabov2007image}}}&
                % \multicolumn{1}{c}{(b) Observations, FID=298} &
                {\makecell[c]{Ambient\\Flow~\cite{kelkar2023ambientflow}}}&
                {\makecell[c]{Ours}} &
                {\makecell[c]{Ground\\Truth}}
                % \multicolumn{1}{c}{(c) AmbientFlow, FID=295} &
                % \multicolumn{1}{c}{(d) Ours, FID=210}
		\end{tabular}
	\end{tabular}
        \caption{\textbf{Posterior samples from the generative model trained on noisy microscopic images.} On the downstream image denoising task, our approach outperforms BM3D~\cite{dabov2007image} and AmbientFlow~\cite{kelkar2023ambientflow} by effectively removing noise while preserving intricate structural details of microscopic images of tubulins.
    % Denoising task on synthetic biological data of tubulins, noise level $\sigma=0.2$. The tubulins are highlighted by yellow fluorescence marks. The images are resized to $64 \time 64$. Due to the complexity of tubulins, our approach outperforms both BM3D and AmbientFlow.
    }
    \label{fig:bio-posterior}
\end{figure*}

\section{Additional posterior sampling results}
\begin{table}[hbt!]
  \centering
  \caption{\textbf{Performance on downstream posterior sampling tasks using MNIST and microscopic images.}
  The optimal results are highlighted in \textbf{bold}.
  % Numerical Results on downstream posterior sampling tasks of MNIST and Synthetic biological data of tubulins. The inputs of MNIST are corrupted with Gaussian noise with $\sigma=0.3$ while the synthetic biological images are corrupted with Gaussian noise with $\sigma=0.2$. Results show that our method is more capable of handling high-resolution and highly corrupted data compared to baselines.
  }
    \begin{tabular}{ccccccc}
    \toprule
    \multirow{2}[4]{*}{\textbf{Method}} & \multicolumn{3}{c}{\textbf{MNIST}} & \multicolumn{3}{c}{\textbf{\makecell[c]{Microscopic images of tubulins}}} \\
\cmidrule{2-7}          & PSNR$\uparrow$  & SSIM$\uparrow$ & LPIPS$\downarrow$   & PSNR$\uparrow$  & SSIM$\uparrow$ & LPIPS$\downarrow$ \\
    \midrule
    Observations & 13.36 & 0.344 & 0.103 & 18.89 & 0.477 & 0.252 \\
    BM3D & 13.57 & 0.427 & 0.088 & 21.28 & 0.542 & \textbf{0.073} \\
    AmbientFlow  & 17.67 & 0.476 & \textbf{0.047} & 15.35 & 0.120 & 0.537\\
    Ours  & \textbf{20.97} & \textbf{0.618} & 0.053 & \textbf{23.33} & \textbf{0.687} & 0.111 \\
    \bottomrule
    \end{tabular}%
  \label{tab:mnisttable}%
\end{table}%

In this section, we present additional results on posterior sampling for the denoising task using diffusion models trained on corrupted MNIST and microscopic images. Table~\ref{tab:mnisttable} demonstrates that our method surpasses both the classical method, BM3D~\cite{dabov2007image}, and the deep learning baseline, AmbientFlow, in terms of PSNR and SSIM. Fig.~\ref{fig:bio-posterior} showcases the posterior samples obtained from denoising problems using generative models trained on noisy microscopic images. Our method achieves superior denoising performance on microscopic images of tubulins, preserving detailed cellular structures visually. These findings underscore the significant potential of our framework in reconstructing clean fluorescent microscopic images, particularly in scenarios where acquiring clean signals is impractical or cost-prohibitive.

\section{Comparison with amortized inference models trained with a clean diffusion prior}
\label{app:clean}

% We provide an additional comparison with flow models trained with clean diffusion prior in this section. The clean diffusion priors are trained with the original image of MNIST and CIFAR-10. We then loaded the pre-trained model using our amortized inference framework. Fig. \ref{fig:clean-prior-comparison} shows the outputs from the conditional flow model and tab. \ref{tab:clean-prior-comparison} provides the numerical results. The results suggest that our method successfully captured the underlying clean data distribution.

\begin{figure*}
	% \vspace*{-0.5cm}
	\centering
	\setlength{\tabcolsep}{1pt}
	\setlength{\fboxrule}{1pt}
	%\vspace*{1.5cm}
	\begin{tabular}{c}
		\begin{tabular}{cccc|cccc}
			% & 
			\small{\makecell[c]{Observation}} & 
                \small{\makecell[c]{Ours}} &
			\small{\makecell[c]{Clean \\Prior}} & 
			\small{\makecell[c]{Ground \\ Truth}}&
                \small{\makecell[c]{Observation}} & 
                \small{\makecell[c]{Ours}} &
			\small{\makecell[c]{Clean \\Prior}} & 
			\small{\makecell[c]{Ground \\ Truth}}\cr
			% \begin{turn}{90} \!\!\! \!\!\! \!\!\! \!\!\! \!\!\! \!\!\!\small{ImageNet} \end{turn} & 
			\multicolumn{1}{c}{
				\begin{overpic}[width=0.120\linewidth]{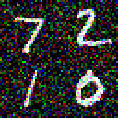}
				\end{overpic}
			}  &
                \multicolumn{1}{c}{
				\begin{overpic}[width=0.120\linewidth]{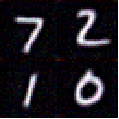}
				\end{overpic}
			}  &
                \multicolumn{1}{c}{
				\begin{overpic}[width=0.120\linewidth]{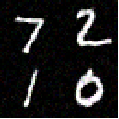}
				\end{overpic}
			}  &
			\multicolumn{1}{c|}{
				\begin{overpic}[width=0.120\linewidth]{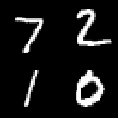}
				\end{overpic}
			}  &
			\multicolumn{1}{c}{
				\begin{overpic}[width=0.120\linewidth]{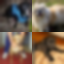}
				\end{overpic}
			}  &
			\multicolumn{1}{c}{
				\begin{overpic}[width=0.120\linewidth]{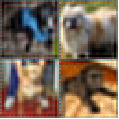}
				\end{overpic}
			}  &
			\multicolumn{1}{c}{
				\begin{overpic}[width=0.120\linewidth]{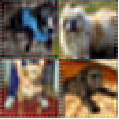}
				\end{overpic}
			}  &
                \multicolumn{1}{c}{
				\begin{overpic}[width=0.120\linewidth]{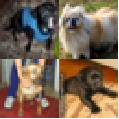}
				\end{overpic}
			} 
			  \\
                \multicolumn{4}{c|}{(a) Amortized Inference Results of MNIST} &
                \multicolumn{4}{c}{(b) Amortized Inference Results of CIFAR-10} 
		\end{tabular}
	\end{tabular}
	\caption{\textbf{Comparative Analysis of Amortized Inference Networks Trained by Our Method and Clean Diffusion Priors.}}
\label{fig:clean-prior-comparison}
% \vspace{-0.1in}
\end{figure*}

We provide an additional comparison of our method to flow models trained with clean diffusion prior. The clean diffusion priors are trained with the clean images of MNIST and CIFAR-10. As illustrated by Fig. \ref{fig:clean-prior-comparison} and Table \ref{tab:clean-prior-comparison}, the performance of the amortized inference networks trained using our method achieves similar results to those trained with clean diffusion priors.

\begin{table}[hbt!]
  \centering
  \caption{\textbf{Amortized inference results for our method and flow trained with clean diffusion prior.}
  % Numerical Results on downstream posterior sampling tasks of MNIST and Synthetic biological data of tubulins. The inputs of MNIST are corrupted with Gaussian noise with $\sigma=0.3$ while the synthetic biological images are corrupted with Gaussian noise with $\sigma=0.2$. Results show that our method is more capable of handling high-resolution and highly corrupted data compared to baselines.
  }
    \begin{tabular}{ccccccc}
    \toprule
    \multirow{2}[4]{*}{\textbf{Method}} & \multicolumn{3}{c}{\textbf{MNIST}} & \multicolumn{3}{c}{\textbf{CIFAR-10}} \\
\cmidrule{2-7}          & PSNR$\uparrow$  & SSIM$\uparrow$ & LPIPS$\downarrow$   & PSNR$\uparrow$  & SSIM$\uparrow$ & LPIPS$\downarrow$ \\
    \midrule
    Observations & 13.57 & 0.210 & 0.591 & 20.91 & 0.582 & 0.182 \\
    Clean Prior  & 24.61 & 0.482 & 0.064 & 22.37 & 0.787 & 0.117\\
    Ours  & 20.73 & 0.399 & 0.160 & 21.97 & 0.787 & 0.135 \\
    \bottomrule
    \end{tabular}%
  \label{tab:clean-prior-comparison}%
\end{table}%

%%%%%%%%%%%%%%%%%%%%%%%%%%%%%%%%%%%%%%%%%%%%%%%%%%%%%%%%%%%%
\clearpage

\end{document}